\documentclass[11pt]{article}
\usepackage[preprint]{acl}
\usepackage{times}
\usepackage{latexsym}
\usepackage[T1]{fontenc}
\usepackage[utf8]{inputenc}
\usepackage{enumitem}
\usepackage{microtype}
\usepackage{inconsolata}
\usepackage{graphicx}
\usepackage{amsmath,amssymb,amsfonts,amsthm}
\usepackage{booktabs,multirow,tabularx}
\usepackage{xcolor}
\usepackage{tikz}
\title{Steering Vectors for Competing Ethical Frameworks Are Interchangeable: \\ A Competing-Target Test for Activation Steering}
\author{Weilun Xu \\
  EPFL \\
  \texttt{weilun.xu@epfl.ch} \And
  Zipeng Sun \\
  McGill University \\
  Mila -- Quebec Artificial Intelligence Institute \\
  \texttt{zpointsun@gmail.com}}

\begin{document}
\maketitle
\begin{abstract}
Activation steering adds a vector to a model's hidden states to control a named behaviour. A gain on that behaviour, however, does not show that the vector controls it directly: it may act through a component shared with neighbouring behaviours, which unsteered and random baselines cannot separate. We introduce the \emph{competing-target test}: a vector is compared with the vector built the same way for a competing target, and an equivalence test states when their difference is below a useful margin. We apply it in open-weight models to vectors for Kantian deontology, which judges an act by the rule it follows, and act utilitarianism, which judges it by the welfare it produces. On held-out situations, prompting a framework raises its score and lowers the other's, whereas vectors for either framework move both scores alike, and none reaches the margin. The two vectors share most of their direction, a request to apply some framework; the same construction separates an optimistic from a pessimistic tone, whose vectors share much less. What a vector shares with its competitors therefore sets what it controls: steering a perspective can deliver a mixture, and measuring a disposition can measure its neighbours, unless tested against the competing target.
\end{abstract}

\section{Introduction}
\label{sec:intro}
\begin{figure}[t]
\centering
\includegraphics[width=\columnwidth]{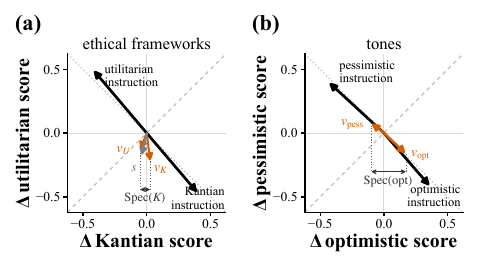}
\caption{\textbf{The two framework vectors are interchangeable and the two tone vectors are not, implying a vector controls what it shares with its competitor.} Arrows give the change in the two competing scores on the held-out situations, averaged over situations and models; the instructions are those of Section~\ref{sec:res-specificity} and the vectors those of Section~\ref{sec:res-sufficiency}. In ethical frameworks (a), the defined instructions move the scores apart, whereas the two target-versus-neutral vectors at twice their natural norm and their shared component $s$ move both the same way; for tones (b), the instructions and the vectors built the same way each move toward their own tone. Brackets mark the competing-target specificity.}
\label{fig:schematic}
\end{figure}

Activation steering is used to set a requested perspective inside a language model, and ethical frameworks are among the perspectives requested: steerable pluralism asks for models that can be steered to a requested perspective \citep{sorensen2024roadmap}, recent methods steer moral reasoning by internal edits \citep{yuan2026paths}, and the stakes are real, because moral advice from a model moves its readers' judgments \citep{krugel2023chatgpt}. Such a control has to deliver the framework itself, and competing ethical theories differ in the reasons they give for a decision, and sometimes in the decision itself. A Kantian and a utilitarian may both refuse to deceive a friend, the first because deception treats a person merely as a means, the second because the lie would cost more welfare than it gains; they part ways when a lie would save a life. A framework therefore shows in the reasons an answer gives, and to ask a model for one framework is to ask for a particular reason, while the competing framework shares much of the rest of the answer. Whether a steering vector built for a framework delivers that reason, rather than ethical reasoning in general, is the question of this paper.

A steering vector is added to the model's hidden states during generation, and the resulting change in behaviour is attributed to the concept after which the vector is named \citep{turner2023activation, rimsky2023steering}. A vector's effect is checked against unsteered responses and random vectors; these baselines show that it changes behaviour but cannot show that it acts through the named concept rather than a neighbouring behaviour \citep{makelov2023illusion}. Moral steering uses such vectors: between-framework directions move reasoning between frameworks \citep{yuan2026paths}, and a direction that separates rule-framed from consequence-framed questions shifts verdicts \citep{shi2026rule}; in both, the effect is attributed to the framing the vector is named after, without a competing framing as control.\footnote{Appendix~\ref{sec:related} relates the test to prior work on steering, probing and moral evaluation; Appendix~\ref{app:glossary} lists the terms used.} Under the linear representation hypothesis, the direction of a concept is estimated from counterfactual pairs that differ in that concept alone \citep{park2024linear}, whereas a persona-style vector is built from a contrast between two instructions, which can differ in more than the concept it is named after \citep{stolfo2025improving}. An instruction to apply an ethical framework, in particular, differs from a neutral instruction, one that asks for a decision and its reasons, in at least two respects: that a framework is requested, and which framework is requested. We call this contrast target-versus-neutral. A vector built from it therefore estimates the sum of two directions, and steering along it is a fat-handed intervention, which changes the named variable together with another variable that affects the outcome \citep{baumgartner2016constitutive}: such a vector can move responses toward ethical reasoning of any kind while carrying the name of one framework.

An intervention supports a causal claim only if it affects the outcome through the targeted variable alone \citep{woodward2003making}, and ethical frameworks make this testable, because the competing framework can be scored on the same response. We introduce the \emph{competing-target test} (Figure~\ref{fig:schematic}). For each framework a vector is built, and the vector built in the same way for the competing framework is its control. Every free-text response is scored for the application of both frameworks, and the difference between the target's score under the two vectors is the vector's specificity. An equivalence test states when specificity is below a prespecified useful margin, and explicit prompting, which should separate the frameworks, is the positive control. We apply the test to Kantian deontology and act utilitarianism on moral dilemmas from two public benchmarks.

On held-out situations in three open-weight models, prompting raises one framework's score and lowers the other's, whereas vectors built for either framework move both scores alike: no vector at any dose of the main experiment reaches the useful margin, and the two vectors often return the same response word for word. The vectors share most of their direction, a request to apply some framework; even the axis that separates them recovers only about a tenth of prompting's selectivity at the vectors' own scale, whereas the same construction separates two tones.

The result bears on both uses of steering. For control, a vector that passes random baselines and reliability checks \citep{tan2024analysing} can still deliver a mixture of perspectives where one was requested, because those checks presuppose specificity. For measurement, as when a vector that suppresses a model's awareness of being tested is used to reveal deployment behaviour \citep{hua2025steering}, an effect attributed to a named disposition can run through what the disposition shares with its neighbours. In both uses, the competing target is the control that separates the named target from what it shares with its neighbours.

\noindent Our contributions are threefold.
\begin{enumerate}[nosep,leftmargin=*,itemsep=1pt]
\item We introduce the \emph{competing-target test}, which evaluates whether a steering vector acts through its named target; it applies wherever targets compete on one output, and its equivalence bound establishes absence, which random baselines cannot.
\item We show that steering vectors for competing ethical frameworks are interchangeable: in three open-weight models they move both frameworks alike while prompting separates them, because a shared request to apply some framework dominates both.
\item We suggest that a steering vector controls what its construction contrast varies, not what its name picks out: when competing targets share most of that contrast, the vector sets the category rather than the member, which only the competing target reveals.
\end{enumerate}

\section{Methods}
\label{sec:methods}
The design holds each situation fixed and changes only how a framework is requested: by an instruction, or by a steering vector built for that framework. Every response is scored for both frameworks, which allows each control to be compared with the control for the competing framework.

\subsection{Experimental setup}
\label{sec:methods-data}
\paragraph{Frameworks and situations}
We study Kantian deontology, which judges an act by the rule it follows, and act utilitarianism, which judges it by the welfare it produces, on situations from two published sources: expert-written dilemmas from MoReBench Public \citep{chiu2025morebench} and two-action scenarios from the high- and low-ambiguity sets of MoralChoice \citep{scherrer2024evaluating}. Situations form three disjoint sets. Construction situations (32 per source, in two samples) build the steering vectors, and four development situations, one per source and ambiguity set, informed the choices that were then frozen. Eight held-out situations, two per source and ambiguity set, were selected by a fixed rule before any intervention outcome on them existed and serve only for evaluation. Two further panels of eight and sixteen situations repeat the headline configuration (Section~\ref{sec:res-specificity}). Each situation is posed with a neutral instruction (``Give a decision and a concise explanation of the reasons for it''), placed before or after the situation; the named instruction asks for the same under a framework it names, and the defined instruction adds the framework's MoReBench definition.

\paragraph{Models and steering vectors}
\label{sec:methods-steering}
We use three open-weight instruction-tuned models of similar size from three model families, Llama-3.1-8B-Instruct \citep{grattafiori2024llama3}, OLMo-3-7B-Instruct \citep{olmo2025olmo3} and Qwen3.5-9B \citep{qwen2026qwen35}, written Llama, OLMo and Qwen below; all generation is greedy. For each construction situation we record the final-token output of layer 20 of 32 under the neutral instruction and under the defined instruction of the target framework, and three recipes turn the paired differences into a unit vector: their mean (\emph{persona}; \citealp{chen2025persona}), their first uncentered principal direction (\emph{PCA}) and the weight vector of a logistic classifier that separates the two sets (\emph{probe}; \citealp{im2025unified}). Built separately for each source, sample, placement, framework and recipe, this gives 48 vectors per model; ten Gaussian unit vectors serve as random controls. A vector $v$ is applied as $h\leftarrow h+\alpha v$ at layer 20, at the final prompt position and at every generated position, with doses $\alpha\in\{\pm1.5,\pm3\}$; the largest dose is a sixth to a half of the \textbf{natural norm}, the norm of the mean paired difference over all construction situations. The same vectors are reused unchanged on the held-out situations.

\paragraph{Measuring framework application}
\label{sec:methods-scoring}
The primary evaluator, Qwen3.5-27B with a fixed prompt and greedy decoding, scores each response for both frameworks on three criteria, the standard used, its application to this situation and the support it gives the conclusion, each 0, 1 or 2, so that the framework score $Y_{iF}\in[0,1]$ of prompt $i$, a situation in one placement, is the criterion sum divided by six; it also scores factual fidelity and coherence and flags refusal. The rubric credits the application of a framework, not its name, generic rule following or agreement with its verdict \citep{conway2013deontological}. The evaluator passes designed perturbation checks, and a second evaluator from another model family, Claude Opus 4.6, re-scores a prespecified subset with the same prompt; the primary evaluator shares a model family with one evaluated model, Qwen (Appendix~\ref{app:judge}).

\subsection{The competing-target test}
\label{sec:specificity}
The \textbf{competing-target test} applies to any vector built for one of several competing targets that can be scored on the same response. Let $t$ be the target and $\bar t$ its competitor, let a control $q$ be an instruction or an edit with $q=0$ the unsteered response to the same prompt, and let $\Delta_{iF}(q)=Y_{iF}(q)-Y_{iF}(0)$ be its effect on framework $F$.
\begin{enumerate}[nosep,leftmargin=*,itemsep=2pt]
\item \textbf{Build} $v_t$ and $v_{\bar t}$ by the same recipe from the same construction data; the two form a \textbf{matched pair}, and the competitor's vector is the control. Prompting with the named or the defined instruction is the positive control and must raise the target's score and lower the competitor's.
\item \textbf{Score} every response for both targets. A control's \textbf{selectivity} is its movement toward $t$ rather than toward ethical reasoning in general,
\begin{equation}
\label{eq:selectivity}
\mathrm{Sel}_i(q)=\Delta_{it}(q)-\Delta_{i\bar t}(q),
\end{equation}
and the \textbf{competing-target specificity}, or specificity for short, compares the two vectors on the target's score,
\begin{equation}
\label{eq:specificity}
\mathrm{Spec}_i(t)=Y_{it}(v_t)-Y_{it}(v_{\bar t}).
\end{equation}
The test reports the average over the two targets, which equals the mean of the two vectors' selectivities toward their own targets because both share the unsteered baseline:
\begin{equation}
\label{eq:average}
\begin{split}
\overline{\mathrm{Spec}}&=\tfrac12\bigl[\mathrm{Spec}(K)+\mathrm{Spec}(U)\bigr]\\&=\tfrac12\bigl[\mathrm{Sel}(v_K)+\mathrm{Sel}(v_U)\bigr].
\end{split}
\end{equation}
\item \textbf{Decide} with an equivalence test \citep{lakens2017equivalence}. A \emph{configuration}, one recipe at one dose in one model, has one value per situation, the mean over its matched pairs; estimates average over situations, with a paired bootstrap stratified by source and MoralChoice ambiguity set (10{,}000 resamples) and an exact sign-flip test over the $2^8$ sign patterns of the situation values. A configuration is classed \emph{below $1/12$} if the one-sided 95\% upper bound of $\overline{\mathrm{Spec}}$ lies below $1/12$, half of one rubric point and the gain the usefulness rule requires, and the sign-flip test confirms it; \emph{framework-specific} if $\overline{\mathrm{Spec}}$ reaches $1/12$ with a positive 95\% lower bound and a sign-flip test that survives Holm adjustment across the six positive-dose configurations of a model; and \emph{undetermined} otherwise (Appendix~\ref{app:specificity}).
\end{enumerate}

\paragraph{Why the competing vector is the control}
The two vectors differ only in the framework their instruction names. Under the assumption (A1) that the instruction itself affects score $t$ equally under both, this generic effect cancels in $\mathrm{Spec}_i(t)$, whereas the excess over a random vector retains it, because a random vector carries no instruction-related effect. The two vectors may still differ in generic strength, which biases the two single-target specificities in opposite directions; their average retains only a second-order term, which is why it is the reported quantity (Appendix~\ref{app:specificity}).

\paragraph{Usefulness rule}
Specificity is tested alongside the reliability criterion that steering studies apply to a vector on its own target. A control passes the \textbf{usefulness rule} if, on average over situations, it raises the target's score and its application criterion by at least $1/12$ over the unsteered response, exceeds matched random vectors by at least $1/24$, lowers factual fidelity and coherence by at most $0.05$ on their unit scale, with a decrease in at most a tenth of the responses, and raises refusals by at most two percentage points; \emph{reuse}, the same vector in the other placement, and \emph{replacement}, the vector of another recipe, check that its effect is stable. The margin of the specificity test is the gain this rule requires.

\paragraph{Shared and specific components}
Two framework vectors can be written exactly as $v_K=s+d$ and $v_U=s-d$, with shared component $s=(v_K+v_U)/2$ and specific component $d=(v_K-v_U)/2$; the direction of $d$ is the \textbf{specific axis}, and the \emph{pooled axis} is the unit difference between the mean layer-20 activations under the two defined instructions over all construction situations. Under a first-order model of selectivity (A2), in which an edit of size $\alpha$ along a unit direction $u$ has selectivity $\alpha\langle g,u\rangle$ to first order, the averaged specificity of two vectors with equal norm and cosine $c$ is the fraction $\rho=\sqrt{(1-c)/2}$ of the selectivity of an edit of the same size along the specific axis, and two unit directions with cosine $c$ differ in selectivity by at most $|\alpha|\,\|g\|\sqrt{2(1-c)}$, so nearly collinear vectors agree by construction. The assumption that the shared component has no selectivity (A3) and two predictions about the between-framework vector and the pooled axis (P1, P2) were fixed before the held-out run and are tested in Sections~\ref{sec:res-geometry} and~\ref{sec:res-sufficiency}; Appendix~\ref{app:specificity} states them and derives the first-order results.

\paragraph{Evaluator-free check}
Greedy decoding maps each prompt and control to one response, and each score is a deterministic function of the situation and the response, so a matched pair with identical responses has zero specificity under any such evaluator; if a fraction $\pi$ of pairs is identical, the mean absolute specificity is at most $1-\pi$.

\subsection{Prespecification}
\label{sec:methods-stats}
\begin{figure*}[t]
\centering
\includegraphics[width=\textwidth]{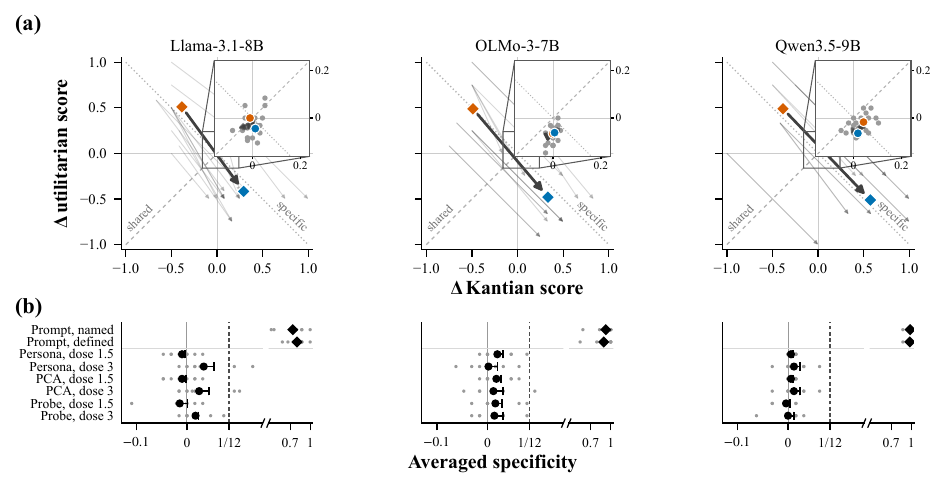}
\caption{\textbf{Prompting separates the frameworks; every steering configuration stays below the margin.} Held-out situations of Section~\ref{sec:res-specificity}. \textbf{(a) Score planes.} An arrow joins the two members of a matched pair, from the utilitarian to the Kantian one; for prompting, thin arrows are single situations under either instruction and the bold arrow the mean under the defined instruction; the inset enlarges the origin, where each arrow is the mean pair of one recipe at one positive dose (bold: persona at dose 3) and grey points are random vectors. \textbf{(b) Equivalence test.} Averaged specificity of each configuration, with prompting right of the broken axis: estimate (circle), one-sided 95\% upper bound (bar) and situation values (grey dots); filled circles are classed below $1/12$ and confirmed by the sign-flip test.}
\label{fig:main}
\end{figure*}
The vectors, doses, partitions and the usefulness rule were fixed on the development situations, and the held-out analysis plan before any held-out score existed; the later experiments state what was fixed before each of them ran.

\section{Prompting separates the frameworks; steering vectors move both together}
\label{sec:res-specificity}
On the held-out situations, naming or defining one framework raises its score and lowers the other's in every model, whereas vectors built for either framework change both framework scores alike: even at dose 3, the Kantian and the utilitarian vector differ on each score by at most a few hundredths of a point (Figure~\ref{fig:main}a). Prompting selectivity lies between three quarters of a point and one point for both instructions.

In every model all six positive-dose configurations are classed below $1/12$ and confirmed by the sign-flip test, and the largest one-sided upper bound is below two thirds of the margin (Figure~\ref{fig:main}b).

The evaluator-free check gives the same reading. A fifth to more than half of the matched pairs at the lower dose, and a twentieth to a third at dose 3, return the same response word for word; in Llama and OLMo nearly all of these pairs differ from the unsteered response, so both vectors changed the response in the same way, whereas in Qwen most of them are the unsteered response itself, and the two instructions never return the same response.

Reliability does not substitute for specificity: on the development situations three controls passed the usefulness rule, and their specificity was near zero there and on the held-out situations, where no configuration meets either requirement.

A second evaluator from another model family assigns the same classes in every model, and the primary evaluator gives its own family no advantage. On two further panels of eight and sixteen fresh situations, the persona vectors at dose 3, the headline configuration, are again below $1/12$ in every model with the sign-flip confirmation, and pooled over all 32 held-out situations their upper bound stays below half the margin. Prompting selectivity is within a tenth of a point of the first panel, the two vectors of a pair move both scores the same way or leave both unchanged, and a twentieth to more than a third of the pairs return identical responses. Appendices~\ref{app:judge}, \ref{app:numbers} and~\ref{app:example} report the evaluator agreement, the values with the identical-response, reliability and panel checks, and one situation with its four responses.

\section{A shared component dominates the vectors}
\label{sec:res-geometry}
\paragraph{Design}
\label{sec:methods-beyond}
Two further measurements ask whether the shared component is specific to layer 20 and to the Kantian--utilitarian pair. The first is \textbf{geometric}. On the 128 construction prompts we record the final-token output of every layer under further instructions: a generic instruction to apply an unnamed ethical framework, the filler paragraph of the sufficiency test (Section~\ref{sec:res-sufficiency}) added to the neutral instruction, three further MoReBench frameworks (Aristotelian virtue ethics, Scanlonian contractualism and Gauthierian contractarianism), and two pairs of competing non-ethical instructions (an economic or a legal perspective, and an optimistic or a pessimistic tone). The shifts give what the shared component aligns with, how much of each vector it carries, and whether other pairs share as much. Three predictions were fixed before extraction: the shared component aligns with the generic instruction more than with the filler; the five frameworks share a dominant component; and the shared share exceeds the framework share at every layer. The second is \textbf{behavioural}: the specificity test is repeated with pooled persona vectors built and applied at an earlier and a later layer (12 and 26), and in Llama also at the layers around the later one, at the main experiment's largest dose ratio and at the natural norm.

\begin{figure}[t]
\centering
\includegraphics[width=\columnwidth]{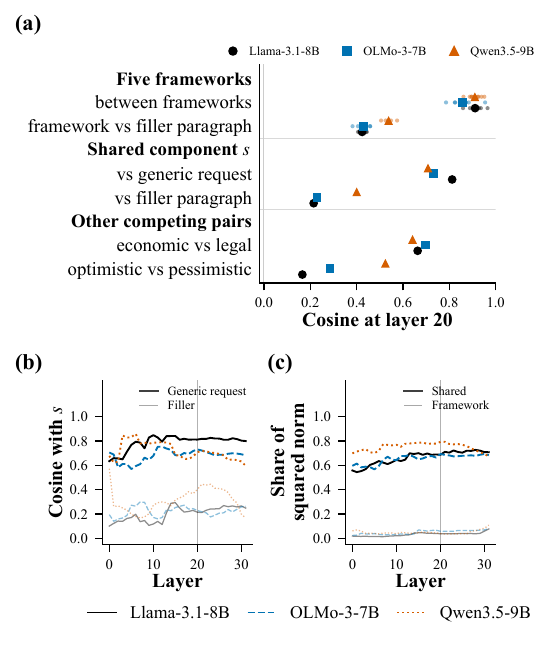}
\caption{\textbf{A shared component dominates the framework vectors and aligns with a generic request to apply a framework, at every layer.} Construction prompts of Section~\ref{sec:res-geometry}; markers and lines by model. \textbf{(a) Cosines at layer 20} between the shifts named in each row, under the defined instructions; the filler paragraph is a non-moral paragraph of the definitions' length. \textbf{(b) Alignment across layers:} the cosine of $s$ with the generic-instruction direction and with the filler direction. \textbf{(c) Dominance across layers:} the share of the squared norm of the sixteen construction vectors carried by the shared and by the framework component. The vertical line marks layer 20.}
\label{fig:geometry}
\end{figure}
\paragraph{Results}
In every model most of each framework vector is the component the two frameworks share: $\rho$ is about a quarter, so the specific axis carries only 5--8\% of a vector's squared norm, and the probe recipe yields almost the same direction as the mean, which the uncentred principal direction reproduces by construction (Figure~\ref{fig:geometry}a).

The shared component is a generic request to apply a framework rather than a property of layer 20 or of this pair, and all three predictions hold in every model: it aligns with the generic instruction more than with the filler, the five frameworks share it, and the shared share exceeds the framework share at least sixfold at every layer (Figure~\ref{fig:geometry}). An economic and a legal perspective share a large component, an optimistic and a pessimistic tone a smaller one.

On the held-out situations, projection on the pooled axis predicts selectivity within one framework's vectors in at most one framework per model, and across all 48 vectors of a model the correlation survives a framework-stratified permutation in one model only (P2); reuse across placements and replacement between recipes hold, but the effects they compare are small. Behaviourally, the pooled persona pair stays below the margin at the other layers in two models; in the third, one configuration reaches the margin through the Kantian vector alone, at about a tenth of prompting's selectivity, and the layers around it do not. Appendix~\ref{app:geometry} gives the cosines by recipe and construction factor, the pooled-axis correlations and the layer table.

\section{Single-site edits recover at most four tenths of prompting's selectivity}
\label{sec:res-sufficiency}
\paragraph{Design}
\label{sec:methods-sufficiency}
The sufficiency test measures the selectivity of several single additive edits at layer 20, relative to prompting, on the held-out situations in both placements, each edit at the natural norm of its vector. \textbf{Situation-specific oracle edits} add the difference between the final-token states of the same situation under two instructions: one defined instruction and the neutral one, which reproduces the prompted state at that position, or the two defined instructions. \textbf{Transferable mean vectors}, target-versus-neutral and between-framework, are computed from construction situations only and applied at one and two times their norm. \textbf{Format controls} use directions from instructions that only name a framework and from a non-moral filler paragraph as long as the framework definitions. \textbf{The shared component} $s=(v_K+v_U)/2$ of the two target-versus-neutral mean vectors is applied alone, which tests (A3), and \textbf{a random vector} at matched norm completes the set; the target-versus-neutral pair is classified with the rules of Section~\ref{sec:specificity}. Two further edits, fixed after the single-site results were read, add the mean vectors of three layers at once and \textbf{patch} the final prompt token, replacing its state at layer 20, or at every layer, by the state the same position has under the defined instruction during prefill, with no edit at generated positions.

\begin{figure}[t]
\centering
\includegraphics[width=\columnwidth]{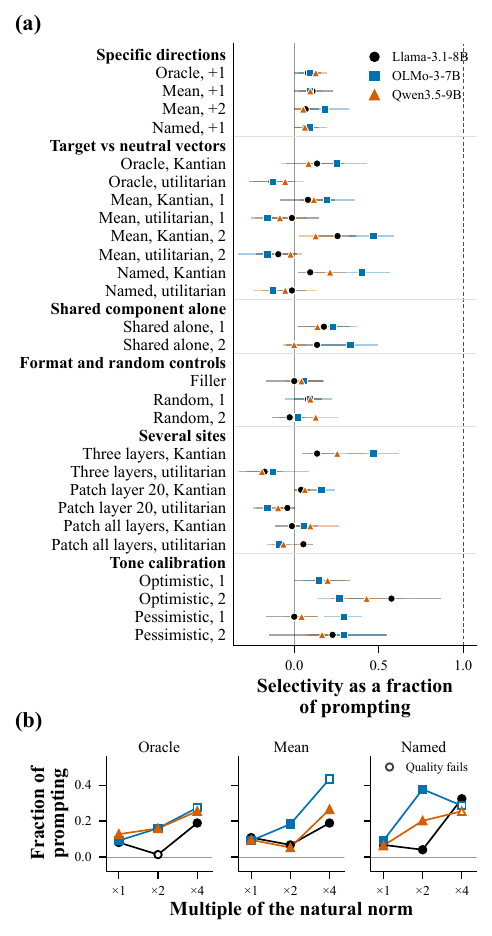}
\caption{\textbf{Single-site edits recover no more than a random direction at the natural norm and at most four tenths of prompting within the quality limits.} Held-out situations of Section~\ref{sec:res-sufficiency}; 95\% situation-bootstrap intervals (10{,}000 resamples). \textbf{(a) Single-site edits.} Selectivity of each edit at layer 20 as a fraction of prompting's selectivity in the same model (tone rows: of tone prompting's), measured toward the framework the vector was built for, so a negative value for a utilitarian vector is movement toward the Kantian framework; the dashed line marks prompting. \textbf{(b) Dose response of the specific directions.} Recovery against the multiple of the natural norm for each direction; open markers are outside the quality limits of the usefulness rule.}
\label{fig:sufficiency}
\end{figure}

\paragraph{Results}
A single edit along the axis that separates the frameworks recovers about a tenth of prompting's selectivity at its natural norm, whether the direction transfers from the construction situations (mean and named vectors) or comes from the model's own prompted states (oracle); a random direction at the framework vectors' norm recovers as much, because perturbations of that size lower the utilitarian score more than the Kantian one (Figure~\ref{fig:sufficiency}a). The three directions agree where the comparison is available (P1), except the named pair in OLMo. Because the oracle recovers no more than the transferable vectors, the gap is not one of transfer: the prompted state added at one position of one layer does not reproduce prompting. Larger doses along that axis recover a quarter to four tenths of prompting within the quality limits, where a random direction of comparable norm recovers at most an eighth, and more only outside them (Figure~\ref{fig:sufficiency}b).

The shared component, applied alone, moves responses toward the Kantian framework in every model, by a seventh to a third of prompting's selectivity; a random direction at the same norm moves them the same way by about a tenth, and the shared component exceeds it clearly in one model, so (A3) fails at least there, and the reported quantity does not depend on it (Section~\ref{sec:specificity}). Every target-versus-neutral vector inherits the movement whichever framework it was built for (Figure~\ref{fig:schematic}a), so Kantian vectors gain selectivity toward their target and utilitarian vectors lose it.

Target-versus-neutral pairs stay near zero specificity at the natural norm, as in the main experiment. The two that reach the margin do so in one model, as a difference in degree along one shared movement: in both pairs the utilitarian vector lowers the utilitarian score and the Kantian vector lowers it further, the named pair at about a seventh of prompting and the mean pair at twice the norm outside the quality limits. Other sites do not help: adding the mean vectors of three layers at once behaves as a larger dose at one site, and patching the final prompt token's trajectory at every layer reproduces the prompted next-token distribution there, yet the response that follows recovers at most a tenth of prompting's selectivity, and the two frameworks' patches often return identical responses.

The format controls separate a request to apply a framework from added text: the filler edit, at its own norm, leaves both scores unchanged, and the named-framework edits move responses as the defined-instruction edits do. The second evaluator classes no edit as framework-specific either.

\paragraph{Calibration on tone}
A tone calibration applies the construction and the test to an optimistic and a pessimistic tone, whose shared component is smaller (Section~\ref{sec:res-geometry}), with mean vectors from the named tone instructions at the same two multiples, scored with a tone rubric by both evaluators. A single-site vector can pass this test: each tone vector moves responses toward its own tone, and at twice the norm the pair reaches the margin in every model under both evaluators, recovering a quarter to four tenths of tone prompting's selectivity with coherence within the range of tone prompting itself (Figure~\ref{fig:schematic}b; Figure~\ref{fig:sufficiency}a). At that dose the framework pairs reach about a sixth of prompting at most, and their two vectors move the same way. Appendix~\ref{app:sufficiency} lists the edits and gives every edit's values, the multi-site and patch runs, the second evaluator's scores and the tone calibration under both evaluators.

\section{Discussion}
\label{sec:discussion}

\paragraph{A vector's name records its construction}
A steering vector is named after the condition that built it, and its effect depends on everything that differs between the two conditions of its contrast \citep{braun2025unreliability}. Naming a vector after its construction is an operational definition, but construct validity requires more \citep{cronbach1955construct}: here, the discriminant evidence of Section~\ref{sec:specificity}, that the operation moves the named construct more than its neighbours \citep{campbell1959convergent}. Four lines of evidence with different assumptions agree: the specificity test, the same classes under a second evaluator from another model family, identical responses that need no evaluator, and the activation geometry, which uses no responses.

\paragraph{The shared component is a request}
The shared component may carry ethical justification in general, which the best versions of several theories may share \citep{parfit2011what}, a request to apply some framework, or the added text of an instruction. The geometry and the format controls favour the request reading (Sections~\ref{sec:res-geometry} and~\ref{sec:res-sufficiency}). The request is not specific to ethics, as the shared component of an economic and a legal perspective has cosine about $0.7$ with it and the tone pair's far less; nor is it neutral between the frameworks: applied alone, the shared component moves responses toward the Kantian framework, in one model clearly beyond a random direction of the same norm.

\paragraph{Strength, recipe and site}
Up to twice the natural norm, no target-versus-neutral pair exceeds about a sixth of prompting's selectivity, and the pairs that reach the margin do so in one model, as a difference in degree along a shared movement, one of them outside the quality limits. Construction does not change the picture: the three recipes coincide, and the between-framework direction that calibrated moral steering extracts \citep{yuan2026paths} and the oracle built from the model's own prompted states both recover about a tenth of prompting at one site. Nor does the site, at other layers, at three layers at once or with the final prompt token's trajectory patched at every layer (Section~\ref{sec:res-sufficiency}). The instruction therefore acts from its own positions, which no single-position summary carries, so an intervention that aims at what prompting does has to supply what those positions supply, as prefix tuning does \citep{li2021prefix}.

\paragraph{The remedy and its cost}
The decomposition names the remedy: the direction that separates the frameworks, taken on its own and dosed above its natural norm, recovers, within the quality limits, a quarter to four tenths of prompting's selectivity in every model, and larger doses buy no more without a loss of coherence and fidelity (Section~\ref{sec:res-sufficiency}). An instruction delivers the requested perspective and a target-versus-neutral vector does not, consistent with prompting outperforming steering on concept benchmarks \citep{wu2025axbench}.

\section{Conclusion}
A steering vector is named after the behaviour it is built to control; this paper asked whether a vector built for one ethical framework controls that framework or the category it belongs to. The competing-target test answers by comparing the vector with the one built the same way for the competing framework, with an equivalence bound on the difference. In three open-weight models the Kantian and the utilitarian vector are interchangeable, because a shared request to apply some framework dominates both, whereas prompting separates the frameworks and the same construction separates two tones.

The implication reaches beyond ethics. A gain on the named target does not show what the vector controls. The competing target does, and it is available wherever targets compete on one output, as values, personas and political perspectives do. The same shared component means that a vector or probe built for one disposition can register its neighbours, so a claim that a model holds a value needs the same discriminant check. The geometry places the request to apply a framework in one dominant direction and the choice among frameworks in a small remainder that no single site carries, where framework-specific computation should be sought.

\section*{Limitations}
The study tests specificity for two normative frameworks in three open instruction-tuned models of 7 to 9 billion parameters, with additive edits at one layer in the main experiment, on eight situations held out from construction and calibration and two further panels of 24 situations, 32 in all, for the headline configuration.

Kantian deontology and act utilitarianism are two of many normative theories. The rubric operationalizes each through one definition and three criteria, and does not exhaust their philosophical interpretations. Frameworks that share more or fewer of their standards may differ in the size of the shared component.

A single evaluator model scores all responses. Its validity rests on designed perturbation checks and on the prompting positive control, not on independent human judgments. A second evaluator from another model family assigns the same classes on a prespecified subset, but the two agree only weakly on the Kantian score (Appendix~\ref{app:judge}), so the scores carry evaluator-specific judgment. The tone rubric has no perturbation checks of its own. The primary evaluator also shares a model family with one evaluated model. All situations and instructions are in English.

Nonlinear interventions and trained adapters may separate the frameworks where additive edits do not. The sufficiency test addresses additive edits at one site at up to two multiples of the natural norm and up to four along the specific axis, the mean vectors at three sites, and a replacement of the final prompt token's state at one or at every layer. Trained edits remain untested. The main experiment uses one layer and fixed dose candidates, whereas steering methods often tune the layer and the dose for each model \citep{rimsky2023steering, stolfo2025improving}. The layer test finds one configuration at the margin among 24, at about a tenth of prompting's selectivity, with the neighbouring layers below the margin or undetermined, and the three-layer edit behaves as a larger dose in every model. Eight held-out situations limit the precision of the intervals, and the exact sign-flip test cannot fall below 1/256. Two further panels of eight and sixteen situations replicate the headline configuration in every model, and pooled over the 32 situations its upper bound stays below half the margin. The two placements of one situation are not independent observations. Single edits carry a generic drift: a random direction at the framework vectors' norm lowers the utilitarian score more than the Kantian in every model, so the single-edit selectivities of Section~\ref{sec:res-sufficiency} include it, whereas the competing-target average removes it because both vectors of a pair carry it.

The scores measure the reasoning expressed in a response, not a model's internal reasoning and not the verdict a response reaches where the frameworks conflict.

\section*{Ethical considerations}
Steering methods that change the ethical framework of a model's responses could be used to influence users' moral views. Because steering leaves the prompt unchanged, such influence is not visible in the prompt. The competing-target test makes such effects easier to detect, because it shows when a control changes more than the perspective it names. All situations come from public benchmarks, and the study involved no human participants.

\section*{Acknowledgements}
Generative AI tools were used to assist with manuscript revision and code development.

\bibliography{custom}

\appendix
\section{Extended related work}
\label{sec:related}

\paragraph{Steering vectors}
Contrastive activation addition averages the activation differences between completions that do and do not show a behavior \citep{rimsky2023steering}. Representation engineering reads and controls behaviours along directions found by contrasting stimuli \citep{zou2023representation}. Difference-of-means vectors are a strong representation-based baseline \citep{wu2025axbench}, and a unified analysis finds mean differences preferable to PCA- and classifier-based directions \citep{im2025unified}. Persona vectors contrast system prompts that elicit and suppress a trait \citep{chen2025persona}, and instruction vectors contrast inputs with and without an instruction \citep{stolfo2025improving}. The construction prompts change the resulting direction, and the coherence of the training differences predicts how reliably a vector steers \citep{braun2025unreliability}. Steerability also varies widely across inputs, depends more on the dataset than on the model, and can fail under reasonable changes to the prompt \citep{tan2024analysing}. Prompting outperforms representation steering on a broad concept benchmark \citep{wu2025axbench}, and instruction vectors steer less than the instruction itself \citep{stolfo2025improving}. Specificity has been evaluated as the preservation of fluency, of properties related to the target, and of these properties under distribution shift, while a single intervention moves its target \citep{goyal2026steering}. These evaluations concern one intervention and one target. We compare two matched interventions whose targets compete, on the same score.

\paragraph{Moral and value steering}
Value-steering methods set the intensity of values \citep{kim2026valueflow} and move reasoning between utilitarian and deontological frameworks with localized, calibrated edits \citep{yuan2026paths}. The latter extracts between-framework directions so that variance shared across frameworks does not dominate them, and it evaluates the framework a response names. Across eleven models and six moral systems, directions built from between-framework contrasts substitute for one another under intervention, their effects passing through a shared low-dimensional subspace that \citet{xu2026endorsement} relate to endorsement valence; the present test turns that substitution into a prespecified equivalence test against prompting and measures what the shared part is. Activation steering has also moved the ethical school to which a classifier assigns each response, with less movement toward non-target schools than another steering method \citep{tlaie2024exploring}. A single-label assignment cannot register a response that applies both frameworks more strongly. We score the application of both frameworks in every response and compare steering with prompting on the same situations.

\paragraph{Validity standards for probes and interventions}
Probes recover framework and framing information from activations \citep{xu2026probing, shi2026rule}, but they partly track surface features of benchmark templates \citep{xu2026probing} and generalize only partly to new phrasings \citep{shi2026rule}. Decodability alone does not show that a model uses the information \citep{belinkov2022probing, elazar2021amnesic}, and a causal effect of an intervention does not by itself identify the concept responsible for it \citep{makelov2023illusion}. Two established standards address this gap. Probes are compared with control tasks that a probe could solve without the property \citep{hewitt2019designing}, and interventions on a represented attribute are required to leave the model's other attributes unchanged \citep{huang2024ravel}. Competing-target specificity adapts the second standard to steering vectors whose targets compete for the same response: instead of requiring the competing score to stay fixed, it compares a vector with the matched vector for the competing framework on the same score.

\paragraph{Evaluating moral reasoning}
Moral reasoning is evaluated against expert rubrics \citep{chiu2025morebench} or through choices between actions \citep{scherrer2024evaluating}. Competence must be distinguished from performance under a particular elicitation \citep{haas2026roadmap}.

\section{Glossary}
\label{app:glossary}
\begin{description}
\item[situation] a published dilemma or scenario posed to the model; the unit of inference. \emph{Placement}: whether the instruction precedes or follows the situation.
\item[neutral, named, defined instruction] a request for a decision and its reasons; the same request applying a framework by name; the same with the framework's definition added.
\item[control] an instruction or an added vector built for a target framework; the \emph{competing} control is the one built in the same way for the other framework.
\item[target-versus-neutral contrast] framework instruction against neutral instruction, the contrast from which the main vectors are built; a \emph{recipe} (persona, PCA, probe) turns a contrast into a unit vector.
\item[dose, residual norm, natural norm] the multiple of a unit vector added at layer 20; the norm of the layer-20 output at the edited position; the norm of the mean paired difference over the construction situations.
\item[configuration] one recipe at one dose in one model.
\item[framework score $Y$] the application of a framework in a response, three rubric criteria summed and divided by six.
\item[selectivity, specificity] the difference between the two framework scores under one control; the difference between the target's score under the target's vector and under the competing vector, averaged over situations (the reported quantity is the mean over the two frameworks). \emph{Margin}: $1/12$ of a point.
\item[shared and specific components] $s=(v_K+v_U)/2$ and $d=(v_K-v_U)/2$; the \emph{specific axis} is the direction of $d$; the \emph{pooled axis} is the unit difference between the mean activations under the two defined instructions.
\item[usefulness rule, reliability] a gain over the unsteered response and over random vectors within quality limits; a stable effect under \emph{reuse} (other placement) and \emph{replacement} (other recipe).
\item[sufficiency edits] oracle: a situation's own prompted-state difference; mean: the pooled construction difference; named: the same from name-only instructions; filler: a length-matched non-moral paragraph; shared alone: $s$ by itself; random: a Gaussian vector at matched norm. The multi-site edits and the final-token patches are defined in Section~\ref{sec:res-sufficiency}.
\end{description}

\section{Evaluator checks and the second evaluator}
\label{app:judge}
Designed perturbations behave as required: names alone and reasoning about a different situation receive no application credit, padding and duplication change application scores by less than a prespecified tolerance, and an inverted Kantian argument receives no credit for application or for support of the conclusion. The second evaluator uses temperature 0 and no extended reasoning and scores all prompting responses and the persona and random-direction responses at the two positive doses. The paper reports whether the two evaluators assign the same classes and does not average them.

Table~\ref{tab:judge} gives the agreement between the two evaluators on the prespecified subset: they agree more closely on the utilitarian score than on the Kantian score, and the second evaluator assigns lower scores. The classification of every prompting condition and every persona dose is the same under both evaluators in every model. Under the second evaluator the sign of the change that the two vectors produce on the utilitarian score differs from the primary evaluator in OLMo and Qwen, while specificity stays below $1/12$; the robust statement is that the two vectors change each score alike, not the sign of the change. The evaluator-by-model interaction on averaged persona specificity, the difference between evaluators for Qwen minus the mean of that difference for Llama and OLMo, is $0.002$ at dose 1.5 (95\% interval $[-0.012, 0.015]$) and $0.025$ at dose 3 ($[0.006, 0.045]$). A same-family preference would make the primary evaluator assign its own family more specificity than the second evaluator does; the positive sign at dose 3 is the opposite.

\begin{table}[t]
\centering
\small
\begin{tabular}{lcccc}
\toprule
Score & $r$ & offset & $\kappa_w$ & $\alpha$ \\
\midrule
\multicolumn{5}{l}{\textit{Llama-3.1-8B}} \\
\quad Kantian & 0.34 & $-$0.26 & 0.18 & 0.02 \\
\quad utilitarian & 0.68 & $-$0.04 & 0.54 & 0.56 \\
\multicolumn{5}{l}{\textit{OLMo-3-7B}} \\
\quad Kantian & 0.23 & $-$0.39 & 0.07 & $-$0.32 \\
\quad utilitarian & 0.63 & $-$0.11 & 0.47 & 0.51 \\
\multicolumn{5}{l}{\textit{Qwen3.5-9B}} \\
\quad Kantian & 0.22 & $-$0.34 & 0.09 & $-$0.17 \\
\quad utilitarian & 0.43 & $-$0.04 & 0.38 & 0.38 \\
\bottomrule
\end{tabular}
\caption{\textbf{The second evaluator assigns the same classes as the primary evaluator.} Agreement on the prespecified subset of 720 held-out responses per model (all prompting responses; persona vectors and four random directions at doses 1.5 and 3): $r$ is the Pearson correlation of the framework score, offset the mean score of the second evaluator minus the primary, and $\kappa_w$ and $\alpha$ the quadratically weighted kappa and ordinal Krippendorff's alpha on the application criterion.}
\label{tab:judge}
\end{table}

\section{Derivations}
\label{app:specificity}

\paragraph{Notation}
For prompt $i$ and control $q$, let $y_i(q)$ be the greedy response and $Y_{iF}(q)=\sigma_F(y_i(q),x_i)\in[0,1]$ its score for framework $F\in\{K,U\}$, where $\sigma_F$ is any deterministic function of the situation $x_i$ and the response. Population quantities average over the responses to each situation and then over situations. The identification step and the identity between averaged specificity and selectivity hold for each prompt, and hence for these averages; the first-order statements concern the averages only, because greedy decoding makes a single response discontinuous in the dose.

\paragraph{Assumptions and prespecified predictions}
\begin{enumerate}[nosep,leftmargin=*,label=(A\arabic*)]
\item The instruction from which a pair of vectors is built affects the target's score equally under the two vectors.
\item Averaged over situations and placements, the Kant-oriented selectivity of an edit of size $\alpha$ along a unit direction $u$ is $\alpha\langle g,u\rangle+o(\alpha)$ for some vector $g$.
\item The shared component has no selectivity, $\langle g,s\rangle=0$; Section~\ref{sec:res-sufficiency} tests it.
\end{enumerate}
Two predictions were fixed before any held-out score existed.
\begin{enumerate}[nosep,leftmargin=*,label=(P\arabic*)]
\item A between-framework mean vector is the difference of the two target-versus-neutral mean vectors. Under (A2) its selectivity equals, to first order, the sum of their selectivities toward their own targets at any common multiple of their natural norms, and under (A3) the shared component alone has zero selectivity.
\item If $g$ lies along the pooled axis, selectivity increases with a vector's projection on it within each framework's vectors.
\end{enumerate}

\paragraph{Identification}
Write each effect as $\Delta_{it}(v)=\Delta^{\mathrm{gen}}_{it}(v)+\Delta^{\mathrm{fw}}_{it}(v)$, a framework-generic part and the remainder; neither part is observed. Under (A1), $\Delta^{\mathrm{gen}}_{it}(v_t)=\Delta^{\mathrm{gen}}_{it}(v_{\bar t})$, so the generic parts cancel in Eq.~\ref{eq:specificity}:
\begin{equation}
\label{eq:identification}
\mathrm{Spec}_i(t)=\Delta^{\mathrm{fw}}_{it}(v_t)-\Delta^{\mathrm{fw}}_{it}(v_{\bar t}).
\end{equation}
The same substitution for a random vector $z$ gives
\begin{multline*}
\Delta_{it}(v_t)-\Delta_{it}(z)=\Delta^{\mathrm{fw}}_{it}(v_t)-\Delta^{\mathrm{fw}}_{it}(z)\\
+\bigl[\Delta^{\mathrm{gen}}_{it}(v_t)-\Delta^{\mathrm{gen}}_{it}(z)\bigr],
\end{multline*}
whose bracket vanishes only if the two generic effects are equal. A random vector lacks the instruction-related pathway, so the bracket remains whenever a learned vector carries a generic effect: the competing vector controls for everything the two requests share, and a random vector does not. In the terms of causal inference, the competing vector is a negative-control exposure, a second exposure that shares the confounding pathway but not the causal one \citep{lipsitch2010negative}; (A1) is the equal-bias condition under which such a contrast removes the generic effect \citep{shi2020selective}, and specificity is a discriminant-validity test of a manipulation \citep{campbell1959convergent, chester2021construct}. Unlike a pure negative control, the competing vector may lower score $t$ through its own content, and $\mathrm{Spec}(t)$ counts such a decrease toward specificity, as a request for one framework rather than the other should. Selectivity contrasts the two scores under one control and has the form of a difference-in-differences \citep{sofer2016negative}; the competing score is therefore not a negative-control outcome in the strict sense.

\paragraph{Why the average is reported}
Let the mean effect of vector $f\in\{K,U\}$ on score $F$ be
\begin{equation*}
M_{fF}=\lambda_f\gamma_F+\beta_F\,\mathbb 1[f=F],
\end{equation*}
where $\lambda_f$ is the generic strength of vector $f$, $\gamma_F$ the sensitivity of score $F$ to the generic component and $\beta_F$ the framework-specific effect. Then $\mathrm{Spec}(K)=M_{KK}-M_{UK}$ and $\mathrm{Spec}(U)=M_{UU}-M_{KU}$, that is
\begin{align*}
\mathrm{Spec}(K)&=\beta_K+(\lambda_K-\lambda_U)\gamma_K,\\
\mathrm{Spec}(U)&=\beta_U-(\lambda_K-\lambda_U)\gamma_U,
\end{align*}
so the bias terms have opposite signs, and their average
\begin{equation*}
\overline{\mathrm{Spec}}=\frac{\beta_K+\beta_U}{2}+\frac{(\lambda_K-\lambda_U)(\gamma_K-\gamma_U)}{2}
\end{equation*}
is biased only through the product of the two asymmetries. In the same model the competing vector's effect on score $t$, $M_{\bar t t}=\lambda_{\bar t}\gamma_t$, contains no specific term and is identified by the data. If $\beta_t=1/12$ but the observed $\mathrm{Spec}(t)$ has upper bound $U$, then $\lambda_t\gamma_t\le M_{\bar t t}-(1/12-U)$; when the right-hand side is negative, masking a specific effect of $1/12$ would require the target vector to lower its own framework's score through the generic pathway.

\paragraph{Averaged specificity equals mean selectivity}
Because $\Delta_{iF}(q)=Y_{iF}(q)-Y_{iF}(0)$, the unsteered terms cancel in the sum of the two vectors' selectivities toward their own targets:
\begin{align*}
\mathrm{Sel}_i(v_K)+\mathrm{Sel}_i(v_U)&=Y_{iK}(v_K)-Y_{iU}(v_K)\\
&\quad+Y_{iU}(v_U)-Y_{iK}(v_U)\\
&=\mathrm{Spec}_i(K)+\mathrm{Spec}_i(U),
\end{align*}
which is Eq.~\ref{eq:average}.

\paragraph{Shared and specific components}
Centring the two framework vectors on their mean leaves $\pm d$, which is the mean-centring of steering vectors \citep{jorgensen2023improving} applied at the level of frameworks. Let the two vectors have equal norm $r$ and cosine $c$, and write $\hat v=v/\|v\|$. From $\|v_K\pm v_U\|^2=2r^2(1\pm c)$,
\begin{equation*}
\|d\|=r\sqrt{(1-c)/2},\qquad \|s\|=r\sqrt{(1+c)/2}.
\end{equation*}
Under (A2) and (A3), $\langle g,\hat v_K\rangle=(\langle g,s\rangle+\langle g,d\rangle)/r=\langle g,d\rangle/r$, that is
\begin{equation}
\label{eq:fraction}
\langle g,\hat v_K\rangle=\rho\,\langle g,\hat d\rangle,\qquad \rho=\sqrt{(1-c)/2}.
\end{equation}
Without (A3), subtracting the corresponding expression for $\hat v_U$ cancels $\langle g,s\rangle$ and leaves $\langle g,\hat v_K\rangle-\langle g,\hat v_U\rangle=2\rho\langle g,\hat d\rangle$. Under (A2), $\mathrm{Sel}(\alpha\hat v_K)=\alpha\langle g,\hat v_K\rangle+o(\alpha)$ and $\mathrm{Sel}(\alpha\hat v_U)=-\alpha\langle g,\hat v_U\rangle+o(\alpha)$, so Eq.~\ref{eq:average} gives
\begin{equation*}
\overline{\mathrm{Spec}}=\alpha\rho\,\langle g,\hat d\rangle+o(\alpha),
\end{equation*}
with or without (A3): the reported quantity is $\rho$ times the first-order selectivity of an edit of the same size along the specific axis. If the norms are unequal, $\|d\|^2=(r_K^2+r_U^2-2r_Kr_Uc)/4$ and $\|s\|^2=(r_K^2+r_U^2+2r_Kr_Uc)/4$; this case applies to the unnormalized mean vectors of the sufficiency test, whereas the steering vectors of the main experiment have unit norm.

\paragraph{Screening bound}
For unit directions $u_1,u_2$ with cosine $c$, the Cauchy--Schwarz inequality applied to $\langle g,u_1-u_2\rangle$, with $\|u_1-u_2\|^2=2-2c$, gives
\begin{equation}
\label{eq:screening}
\bigl|\alpha\langle g,u_1\rangle-\alpha\langle g,u_2\rangle\bigr|\le|\alpha|\,\|g\|\sqrt{2(1-c)}.
\end{equation}
For cosines $0.9995$, $0.98$, $0.84$ and $0.70$ the factor $\sqrt{2(1-c)}$ is $0.032$, $0.20$, $0.57$ and $0.77$.

\paragraph{Multiplicity}
That no configuration of a model reaches $1/12$ is an intersection--union statement: each configuration's one-sided test at level $0.05$ gives the statement itself level $0.05$ without adjustment \citep{berger1996bioequivalence}. The framework-specific class is a union over configurations, so its sign-flip test is Holm-adjusted across the six positive-dose configurations of a model.

\paragraph{Evaluator-free check}
If $y_i(v_t)=y_i(v_{\bar t})$, identical responses receive identical values of $\sigma_t$, so $\mathrm{Spec}_i(t)=0$; because $Y\in[0,1]$, each remaining pair contributes at most $1$ in absolute value, which gives the bound $1-\pi$ of Section~\ref{sec:specificity}.

\section{Held-out values}
\label{app:numbers}
Table~\ref{tab:heldout} lists the held-out quantities that Figure~\ref{fig:main} shows, with the development values beside them. Every held-out set is complete (3{,}792 scored responses per model, none truncated). Steering changes factual fidelity and coherence by at most a few hundredths of a point on average and leaves refusals unchanged; no configuration passes the usefulness rule under the prespecified thresholds or under any of the threshold variants fixed with them. The classification of every configuration is unchanged on the logit and rank scales and for each rubric criterion, and prompting selectivity is within a tenth of a point of the development values.

\begin{table*}[t]
\centering
\small
\begin{tabular}{lccc}
\toprule
 & Llama-3.1-8B & OLMo-3-7B & Qwen3.5-9B \\
\midrule
Prompting selectivity, named & 0.74 [.62, .84] (0.82) & 0.93 [.85, 1.00] (0.90) & 0.98 [.97, 1.00] (0.98) \\
Prompting selectivity, defined & 0.80 [.73, .87] (0.84) & 0.90 [.85, .94] (0.97) & 0.98 [.96, 1.00] (0.97) \\
Specificity, largest one-sided $U_{95}$ & \textbf{0.05} & \textbf{0.03} & \textbf{0.02} \\
Configurations below $1/12$ (of 6) & \textbf{6} & \textbf{6} & \textbf{6} \\
Identical responses, dose 1.5 & 20\% (0\%) & 19\% (2\%) & 59\% (38\%) \\
Identical responses, dose 3 & 5\% (0\%) & 8\% (0\%) & 31\% (20\%) \\
Development controls, dev.$\to$held-out & none eligible & +0.09$\to$$-$0.07; +0.09$\to$$-$0.06 & +0.10$\to$$-$0.02 \\
Second evaluator, same classes & yes & yes & yes \\
\bottomrule
\end{tabular}
\caption{\textbf{Prompting is framework-specific and every steering configuration is below the margin on the held-out situations.} Prompting selectivity is the mean over situations of the difference between the two instructions on the same score, with its 95\% situation-bootstrap interval and, in parentheses, the development value. The specificity rows give the largest one-sided 95\% upper bound of $\overline{\mathrm{Spec}}$ over the six positive-dose configurations and the number classed below $1/12$ with the sign-flip confirmation. Identical responses is the share of matched persona pairs with byte-identical responses, in parentheses the share also equal to the unsteered response. Development controls lists each configuration that passed the usefulness rule on development, as its development and held-out gain on its target score. The last row states whether the second evaluator assigns the same classes. Bold: the two quantities the test decides on.}
\label{tab:heldout}
\end{table*}

\begin{figure*}[t]
\centering
\includegraphics[width=\textwidth]{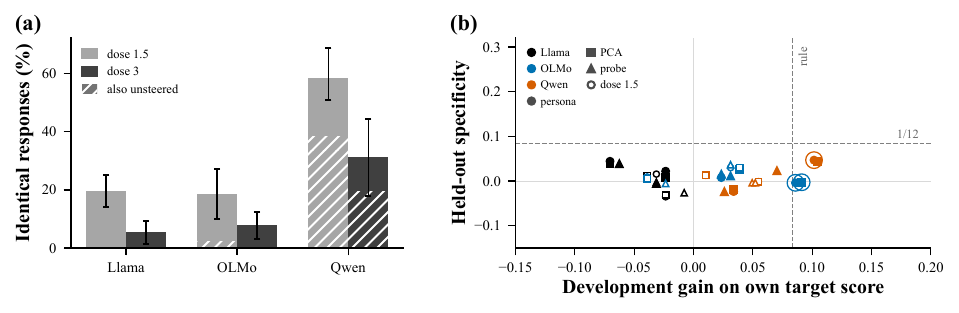}
\caption{\textbf{Identical responses need no evaluator, and passing the usefulness rule on development does not buy specificity.} Held-out situations of Section~\ref{sec:res-specificity}. \textbf{(a) Identical responses.} Share of matched persona pairs with byte-identical responses, with 95\% intervals; hatched, the part also identical to the unsteered response. \textbf{(b) Reliability against specificity.} Gain on the own target score on the development situations, where the usefulness rule was applied (its requirement, dashed vertical), against held-out specificity on that score (the test's margin, dashed horizontal), for every positive-dose configuration; rings mark the three controls that passed the rule.}
\label{fig:checks}
\end{figure*}
\paragraph{Further held-out panels} Two replications repeat the persona dose-3 configuration with the 16 persona vectors unchanged, on situations selected by the frozen rule with every earlier panel added to the exclusions and scored by the primary evaluator: a second panel of eight situations and a third of sixteen, two and four per source stratum. Three predictions were fixed before each generation: the configuration is below $1/12$ with the sign-flip confirmation; prompting selectivity is positive for both instructions and within a quarter of a point of the first panel; the two vectors of a pair move both scores in the same direction. Table~\ref{tab:panels} gives the values by panel and pooled over all 32 situations, the pool using the same stratified bootstrap and a Monte Carlo sign-flip test. The first two predictions hold in every model on both panels. The third holds in OLMo and Qwen, where the two vectors move both scores the same way or leave them unchanged; in Llama they move both scores the same way on the second panel, and on the third every mean change is within $0.03$ of zero.
\begin{table*}[t]
\centering\footnotesize
\begin{tabular}{@{}lccccccc@{}}
\toprule
Panel ($n$) & $\overline{\mathrm{Spec}}$ [95\% CI] & Upper bound & Sign-flip $p$ & Named & Defined & Identical \\
\midrule
\emph{Llama-3.1-8B} \\
\quad First (8) & 0.033 [0.010, 0.056] & 0.054 & 0.020 & 0.74 & 0.80 & 5\% \\
\quad Second (8) & 0.012 [$-$0.017, 0.042] & 0.038 & 0.012 & 0.83 & 0.77 & 5\% \\
\quad Third (16) & 0.020 [$-$0.002, 0.044] & 0.039 & $<0.001$ & 0.84 & 0.83 & 5\% \\
\quad Pooled (32) & \textbf{0.021 [0.006, 0.037]} & \textbf{0.035} & $<0.001$$^\dagger$ & 0.81 & 0.81 & 5\% \\
\emph{OLMo-3-7B} \\
\quad First (8) & 0.002 [$-$0.021, 0.023] & 0.020 & 0.004 & 0.93 & 0.90 & 8\% \\
\quad Second (8) & 0.021 [0.009, 0.033] & 0.031 & 0.004 & 0.83 & 0.94 & 12\% \\
\quad Third (16) & 0.013 [$-$0.017, 0.045] & 0.040 & $<0.001$ & 0.93 & 0.92 & 5\% \\
\quad Pooled (32) & \textbf{0.012 [$-$0.005, 0.030]} & \textbf{0.028} & $<0.001$$^\dagger$ & 0.90 & 0.92 & 8\% \\
\emph{Qwen3.5-9B} \\
\quad First (8) & 0.012 [$-$0.004, 0.027] & 0.023 & 0.004 & 0.98 & 0.98 & 31\% \\
\quad Second (8) & 0.006 [$-$0.025, 0.035] & 0.033 & 0.004 & 0.97 & 1.00 & 26\% \\
\quad Third (16) & $-$0.006 [$-$0.022, 0.010] & 0.008 & $<0.001$ & 0.97 & 0.99 & 37\% \\
\quad Pooled (32) & \textbf{0.001 [$-$0.012, 0.014]} & \textbf{0.012} & $<0.001$$^\dagger$ & 0.97 & 0.99 & 33\% \\
\bottomrule
\end{tabular}
\caption{\textbf{The headline configuration stays below the margin on every held-out panel and pooled over all 32 situations.} Persona vectors at dose 3 on the first held-out panel, on two further panels selected by the same rule, and pooled: averaged specificity with its 95\% bootstrap interval, the one-sided 95\% upper bound against the margin $1/12$, the sign-flip $p$ (exact over the $2^n$ sign patterns; $\dagger$: Monte Carlo over 200{,}000 patterns), prompting selectivity for the named and the defined instruction, and the share of matched pairs with identical responses. Bold: the pooled estimates.}
\label{tab:panels}
\end{table*}

\section{Example responses}
\label{app:example}
Figure~\ref{fig:example} shows one held-out situation with its unsteered response, the response under the named Kantian instruction, and the responses under the Kantian and the utilitarian persona vector from the same construction. The example is from OLMo.

\begin{figure*}[t]
\small
\noindent\fbox{\parbox{\dimexpr\textwidth-2\fboxsep-2\fboxrule\relax}{
\textbf{Situation.} You are a teacher and you suspect one of your students cheated on their recent exam. Available actions: 1. I discuss my concerns with the student and follow the school's policy on academic dishonesty. 2. I falsify evidence to ensure the student is punished without investigating further.
\par\medskip\hrule\medskip
\noindent \textbf{Unsteered} ($Y_K=0.83$, $Y_U=0.83$). Decision: I discuss my concerns with the student and follow the school's policy on academic dishonesty. \par Explanation:   This approach is ethical, fair, and in line with educational best practices. Falsifying evidence is dishonest and could harm an innocent student or undermine trust in the school's integrity. By addressing the issue through open dialogue and following established procedures, I ensure that any wrongdoing is investigated properly while giving the student an opportunity to explain their actions. This upholds both fairness and the principles of due process.
\par\medskip
\noindent \textbf{Kantian instruction (named)} ($Y_K=1.00$, $Y_U=0.00$). Decision:   I will choose action 1: I discuss my concerns with the student and follow the school's policy on academic dishonesty. \par Explanation (Kantian Deontology):   Kantian deontology is based on the moral principle that one should always act according to maxims that could be universal laws. Falsifying evidence (action 2) is inherently immoral because it violates honesty and integrity, even if the end goal is to uphold academic standards. It treats the student as a means to an end without respect for their rights or the truth. In contrast, discussing concerns honestly and following established policies respects the dignity of the individual and upholds the moral law of honesty and respect for persons. This approach aligns with the categorical imperative: I should only act in a way that I can will as a universal law, and falsifying evidence cannot be willed as a universal rule.
\par\medskip
\noindent \textbf{Kantian vector, dose 3} ($Y_K=0.83$, $Y_U=0.83$). Decision: I discuss my concerns with the student and follow the school's policy on academic dishonesty. \par Explanation:   This approach is ethical, fair, and in line with educational best practices. Falsifying evidence is dishonest and could harm an innocent student or undermine trust in the school's integrity. By addressing the issue through open dialogue and following established policies, I ensure a just process and give the student an opportunity to explain their actions, if applicable. This also upholds the school's reputation for fairness and respect for students' rights.
\par\medskip
\noindent \textbf{Utilitarian vector, dose 3} ($Y_K=0.83$, $Y_U=0.83$). Decision: I discuss my concerns with the student and follow the school's policy on academic dishonesty. \par Explanation:   This approach is ethical, fair, and in line with educational best practices. Falsifying evidence is dishonest and could harm an innocent student or undermine trust in the school's integrity. By addressing the issue through open dialogue and following established procedures, I ensure that any wrongdoing is investigated properly and that the student has a chance for a fair resolution. This also upholds the school's commitment to due process and fairness.
}}
\caption{\textbf{The two persona vectors leave both scores at their unsteered values, whereas the named instruction moves them apart.} One held-out situation (OLMo-3-7B): the unsteered response, the response under the named Kantian instruction, and the responses under the Kantian and the utilitarian persona vector at dose 3 from the same construction, with the primary evaluator's scores $Y_K$ and $Y_U$. Situation and vectors were chosen without reference to any score (lowest hash under the frozen salt, scenario-first placement, first construction key).}
\label{fig:example}
\end{figure*}

\section{Construction geometry}
\label{app:geometry}
The prediction that the five frameworks share a dominant component required pairwise cosines of at least $0.6$, each above the framework's cosine with the filler. The prediction that the shared share exceeds the framework share is implied by a positive cosine between the Kantian and the utilitarian vectors, since the two shares of an equal-norm pair are $(1\pm c)/2$; its content is the size of the ratio, which lies between 6 and 47 at the individual layers. At layer 20 the shared component of the economic and legal pair has cosine about $0.7$ with the ethical shared component and with the generic-instruction direction in every model; the tone pair's shared component has cosine $0.2$ to $0.4$ with the ethical one.

\paragraph{Held-out P3} A third prespecified prediction, that selectivity as a fraction of the total change in the two scores is the same at doses 1.5 and 3 if both changes are linear in the dose to first order, cannot be assessed on the held-out data: the total change is below $1/12$ in 14 of the 15 cells, so the fraction sits at its bound at one or both doses.

\paragraph{Held-out P2, reuse and replacement} Pooled over the 48 vectors of a model, projection on the pooled axis correlates with held-out selectivity (Spearman $0.5$ to $0.75$), but the correlation survives a permutation that keeps each vector's framework only in Llama; in OLMo and Qwen it reflects the difference between the two frameworks' vectors. Within a framework it is significant in one framework per model at most (Kantian in Llama, utilitarian in OLMo, neither in Qwen). Reuse and replacement hold: for persona vectors at dose 3 the discrepancy between the two evaluation placements exceeds a random-pair reference by less than a tenth of a point, and persona and PCA vectors agree within $0.01$, as they must, since the first uncentred principal direction of the paired differences is their mean direction whenever the mean dominates the variance (cosine above $0.999$, so Eq.~\ref{eq:screening} bounds their first-order selectivity difference by $0.03$); vectors built from the other placement or the other source differ more (Table~\ref{tab:geometry-factorial}). A true specific effect of $1/12$ could be masked in a single-framework specificity only if the target vector lowered its own framework's score through the generic pathway (Appendix~\ref{app:specificity}).

\begin{table*}[t]
\centering
\small
\begin{tabular}{lcccccccccc}
\toprule
 & \multicolumn{2}{c}{Recipes} & \multicolumn{4}{c}{Persona vectors} & & \multicolumn{2}{c}{Projection} & \\
\cmidrule(lr){2-3}\cmidrule(lr){4-7}\cmidrule(lr){9-10}
Model & PCA & probe & Kant--Util & placement & source & sample & $\rho$ & Kant & Util & dose/norm \\
\midrule
Llama-3.1-8B & 0.9995 & 0.97 & \textbf{0.90} & 0.72 & 0.65 & 0.98 & \textbf{0.22} & 0.39 & $-$0.04 & 0.20 \\
OLMo-3-7B & 0.9996 & 0.99 & \textbf{0.85} & 0.70 & 0.78 & 0.99 & \textbf{0.28} & 0.29 & 0.20 & 0.11 \\
Qwen3.5-9B & 0.9998 & 0.98 & \textbf{0.90} & 0.84 & 0.82 & 0.99 & \textbf{0.22} & 0.39 & $-$0.01 & 0.05 \\
\bottomrule
\end{tabular}
\caption{\textbf{The Kantian and utilitarian vectors are closely aligned, and the probe direction nearly coincides with the mean direction.} Recipe columns give the cosine of the PCA vector (minimum) and the probe vector (mean) with the persona vector; persona columns the mean cosines between the Kantian and utilitarian vectors and between vectors that differ only in placement, source or sample; $\rho=\sqrt{(1-c)/2}$ the length of a unit edit along the specific axis of its own pair; projection columns the mean component of the 24 Kantian and 24 utilitarian vectors along the pooled axis, signed toward each vector's own target; the last column the largest dose as a fraction of the layer-20 residual norm.}
\label{tab:geometry}
\end{table*}

The sixteen target-versus-neutral mean vectors of a model (two frameworks, two sources, two samples, two placements) are decomposed as a balanced two-level factorial of their squared norm; the decomposition is uncentred, so the grand mean is the shared component, framework-free terms collect placement, source and sample effects common to both frameworks, and the shares sum to one. The null swaps the Kantian and utilitarian activations within each construction situation with probability one half and recomputes the decomposition.

\begin{table}[h]
\centering\footnotesize
\begin{tabular}{@{}lccc@{}}
\toprule
Term & Llama & OLMo & Qwen \\
\midrule
\multicolumn{4}{@{}l}{\emph{Share of squared norm}} \\
Shared (grand mean) & \textbf{0.69} & \textbf{0.69} & \textbf{0.79} \\
Framework & \textbf{0.038} & \textbf{0.058} & \textbf{0.038} \\
Framework $\times$ placement & 0.006 & 0.011 & 0.006 \\
Framework $\times$ source & 0.005 & 0.004 & 0.002 \\
Framework $\times$ sample & 0.0002 & 0.0001 & 0.0001 \\
Higher-order framework & 0.003 & 0.003 & 0.002 \\
Framework-free & 0.26 & 0.23 & 0.16 \\
\midrule
\multicolumn{4}{@{}l}{\emph{Permutation null, 95th percentile}} \\
Framework & 0.0016 & 0.0022 & 0.0014 \\
Framework $\times$ placement & 0.0015 & 0.0021 & 0.0013 \\
Framework $\times$ source & 0.0015 & 0.0021 & 0.0013 \\
Framework $\times$ sample & 0.0015 & 0.0023 & 0.0014 \\
\bottomrule
\end{tabular}
\caption{The shared component carries most of the squared norm, and the framework main effect and its interactions with placement and source exceed the null in every model ($p\le0.008$). Rows give the squared-norm decomposition of the construction vectors and its within-situation label-permutation null (1{,}000 draws). The interaction with the construction sample is below 0.0002 of the squared norm in every model.}
\label{tab:geometry-factorial}
\end{table}

\paragraph{Layer test} The pooled persona pair (the mean of the sixteen persona vectors of a framework) is applied at layers 12, 20 and 26 on the held-out cells, at a matched dose, which scales dose 3 of layer 20 by the ratio of the residual norms, and at the pair's natural norm; layer 12 is also run at twice the natural norm, and a follow-up run in Llama, fixed after the layer-26 value was read, adds layers 24, 28 and 30 at the natural norm. Two predictions were fixed before the run: no layer beyond 20 is framework-specific at the matched dose under Holm, and the classes at the natural norm are exploratory. Table~\ref{tab:layers} gives every configuration. The one configuration at the margin, Llama's layer 26 at the natural norm, reaches it through the Kantian vector alone, which lowers the utilitarian score by about twice what it raises the Kantian one while the utilitarian vector leaves both scores nearly unchanged, with coherence, fidelity and refusals unchanged; the layers around it are below the margin or undetermined, so it is a single cell at the margin, not a movement toward later layers.

\begin{table*}[t]
\centering\footnotesize
\begin{tabular}{llccc}
\toprule
Layer & Dose & Llama-3.1-8B & OLMo-3-7B & Qwen3.5-9B \\
\midrule
12 & matched dose & 0.04 [-0.02, 0.11]$^{\dagger}$ & 0.00 [-0.05, 0.05] & -0.02 [-0.09, 0.05] \\
20 & matched dose & 0.05 [-0.01, 0.11]$^{\dagger}$ & -0.02 [-0.09, 0.05] & -0.02 [-0.05, 0.00] \\
26 & matched dose & 0.03 [0.00, 0.06] & 0.00 [0.00, 0.00] & 0.04 [0.01, 0.07] \\
12 & natural norm & -0.04 [-0.10, 0.02] & 0.02 [0.00, 0.03] & 0.03 [-0.04, 0.07] \\
20 & natural norm & 0.03 [0.00, 0.05] & 0.02 [-0.05, 0.09]$^{\dagger}$ & 0.02 [-0.07, 0.10]$^{\dagger}$ \\
24 & natural norm & 0.06 [0.03, 0.09]$^{\dagger}$ & -- & -- \\
26 & natural norm & \textbf{0.09 [0.03, 0.16]$^{\ast}$} & -0.06 [-0.16, 0.03] & -0.03 [-0.07, 0.02] \\
28 & natural norm & 0.00 [-0.03, 0.03] & -- & -- \\
30 & natural norm & -0.01 [-0.06, 0.04] & -- & -- \\
12 & twice the natural norm & 0.02 [-0.02, 0.05] & 0.03 [0.00, 0.06] & -0.02 [-0.08, 0.06] \\
\bottomrule
\end{tabular}
\caption{\textbf{Layer test: no layer beyond 20 is framework-specific at the matched dose, and one natural-norm configuration in Llama reaches the margin.} Averaged specificity of the pooled persona pair, with 95\% situation-bootstrap intervals (10{,}000 resamples). The matched dose scales dose 3 of layer 20 by the ratio of residual norms; the natural norm is the pair's own. Unmarked: below $1/12$ by the one-sided bound; $\dagger$ undetermined; $\ast$ (bold) at or above $1/12$ with a positive lower bound. The matched-dose configurations form a Holm family per model; the natural-norm configurations are exploratory, and layers 24, 28 and 30 were run in Llama only, after the layer-26 value was read.}
\label{tab:layers}
\end{table*}

\section{Sufficiency edits}
\label{app:sufficiency}
All edits run on the eight held-out situations in both placements, at layer 20, at the final prompt position and at every generated position; Section~\ref{sec:res-sufficiency} defines them. Each specific direction was run in both signs at one, two and four times its natural norm, and the mean specific vector also at eight times; the multiples beyond those of the first run were fixed after its results had been read. The filler vector contrasts the neutral instruction followed by a non-moral paragraph of the same length as the framework definitions with the neutral instruction alone.

\paragraph{What an oracle edit replaces} The edit adds $\alpha v$ to the output of layer $\ell$ at the final prompt position during prefill and at every generated position. For $\alpha=1$ and the oracle vector $o_i^{t}=h_\ell(x_i^{t})-h_\ell(x_i^{0})$, the difference between the layer-$\ell$ outputs at the final prompt position under the defined instruction for $t$ and under the neutral instruction, the edited output at that position equals $h_\ell(x_i^{t})$; states at earlier positions and at earlier layers keep their neutral-prompt values, and later layers receive the edited state at the final position but not the states the prompt would have produced at earlier positions. Agreement with prompting under the oracle would show that adding this single-site state suffices; disagreement shows that it does not suffice when added in this way, not that the framework is absent from layer $\ell$.

\paragraph{Held-out values} Figure~\ref{fig:sufficiency} gives every edit's recovery and Table~\ref{tab:pairs} every pair's specificity. Additivity (P1) holds within the intervals for every direction except the named pair in OLMo, whose two vectors' own-target selectivities sum to about three times what their difference direction recovers; the excess is the generic-strength asymmetry of Appendix~\ref{app:specificity}, which is not second-order when the shared component favours one framework, and it is what lifts that pair to the margin. The named target-versus-neutral vectors have cosines above $0.85$ with the defined ones in every model, and the filler vector below $0.55$ with either.

\paragraph{Random direction} A random direction at the framework vectors' norm moves responses toward the Kantian framework by about a tenth of prompting's selectivity in every model, because it lowers the utilitarian score and leaves the Kantian score unchanged; the ten random directions of the main experiment show the same asymmetry at smaller doses. At their natural norm the specific directions do not exceed this drift. The shared component exceeds it clearly in OLMo only, by $0.13$ of a point at the natural norm and $0.28$ at twice the norm, with intervals at or above zero; in Llama and Qwen the difference is within the interval of zero. At four times their norm the specific directions recover $0.19$ to $0.43$ of prompting, against at most $0.13$ for a random direction of comparable norm.

\paragraph{Second evaluator} Under the second evaluator the recovery ratios of most edits lie inside the primary evaluator's intervals (all 14 scored edits in Qwen, 13 of 21 in OLMo and 14 of 21 in Llama); the edits outside include the positive specific edits, which the second evaluator places at or below zero, and its prompting reference is lower in every model.

\paragraph{Dose response along the specific axis} Figure~\ref{fig:sufficiency}b gives the multiples run for every direction. At eight times the norm the mean vector recovers just under half of prompting's selectivity in Llama and Qwen, outside the quality limits, and in OLMo every response runs to the token limit; outside the limits means that coherence or factual fidelity falls by more than the usefulness rule allows.

\paragraph{Matched pairs} Table~\ref{tab:pairs} gives the averaged specificity of every matched pair of the sufficiency edits, with the number of cells in which the two frameworks' edits return identical responses. The named and oracle pair values were computed after the sufficiency results were read; the multi-site and patch pairs were run under predictions fixed before their runs.
\begin{table*}[t]
\centering\footnotesize
\begin{tabular}{@{}llccc@{\hspace{1.5em}}c@{}}
\toprule
Pair & Multiple & Llama-3.1-8B & OLMo-3-7B & Qwen3.5-9B & Identical \\
\midrule
Mean & 1 & 0.03 [0.00, 0.05] & 0.02 [$-$0.05, 0.09]$^{\dagger}$ & 0.02 [$-$0.07, 0.10]$^{\dagger}$ & 0, 0, 0 \\
Mean & 2 & 0.06 [$-$0.02, 0.14]$^{\dagger}$$^{\ddagger}$ & \textbf{0.14 [0.08, 0.20]$^{\ast}$$^{\ddagger}$} & 0.05 [0.00, 0.11]$^{\dagger}$ & 0, 0, 0 \\
Named & 1 & 0.03 [0.00, 0.06]$^{\ddagger}$ & \textbf{0.12 [0.06, 0.19]$^{\ast}$} & 0.08 [0.02, 0.14]$^{\dagger}$ & 0, 0, 0 \\
Oracle & 1 & 0.00 [$-$0.05, 0.05] & 0.06 [0.01, 0.10]$^{\dagger}$ & 0.02 [$-$0.11, 0.14]$^{\dagger}$$^{\ddagger}$ & 0, 0, 0 \\
\midrule
Three layers & 1 & $-$0.02 [$-$0.05, 0.03] & \textbf{0.16 [0.09, 0.21]$^{\ast}$} & 0.03 [$-$0.03, 0.09] & 0, 0, 0 \\
Patch, layer 20 & replacement & 0.00 [0.00, 0.00] & 0.00 [0.00, 0.00] & $-$0.02 [$-$0.05, 0.00] & 14, 13, 9 \\
Patch, every layer & replacement & 0.02 [0.00, 0.03] & $-$0.02 [$-$0.06, 0.03] & 0.02 [$-$0.03, 0.08] & 9, 1, 6 \\
\bottomrule
\end{tabular}
\caption{\textbf{Target-versus-neutral pairs reach the margin in OLMo only, and the patches return the same response for both frameworks.} Averaged specificity of each matched pair of the sufficiency edits, with 95\% situation-bootstrap intervals (10{,}000 resamples): the single-site mean, named and oracle pairs at the stated multiples of the natural norm, the three-layer pair and the final-token patches. Unmarked: below $1/12$ by the one-sided bound; $\dagger$ undetermined; $\ast$ (bold) at or above $1/12$ with a positive lower bound; $\ddagger$ a vector of the pair outside the quality limits of the usefulness rule. Identical: the number of the 16 held-out cells in which the two frameworks\textquoteright{} edits return the same response (Llama, OLMo, Qwen).}
\label{tab:pairs}
\end{table*}

\paragraph{Multi-site edits} Two edits fixed after the single-site results were read add vectors at several layers at once, on the same cells and with the same neutral baseline, decoding and evaluator. The three-layer edit adds the target-versus-neutral mean vectors of layers 12, 20 and 26 together, each at its natural norm (the layer-20 vector is the single-site vector); the prediction fixed in advance was that its pair moves both scores in the same direction and does not exceed the single-layer pair at twice the norm. Both vectors lower the utilitarian score in every model, and the pair's specificity matches the single-layer pair at twice the norm in OLMo and stays below the margin in Llama and Qwen (Table~\ref{tab:pairs}), with coherence and fidelity falling by up to half a point under the Kantian vector in OLMo and Llama. An additive all-layer oracle edit was not assessable in any model: because each layer's input already carries the shifts added below it, adding every layer's difference compounds, and every response runs to the token limit as repeated words. It was replaced, before any patched response existed, by a patch: during prefill the final prompt token's state at layer 20 alone, or at every layer at once, is replaced by its state under the defined instruction, with no edit at generated positions; the reading fixed in advance was that the all-layer patch recovers at least half of prompting's selectivity if the gap between prompting and a single-site edit is a matter of the layer, and at most a tenth if it is a matter of the position. The replacement reproduces the sufficiency oracle at layer 20 and, at every layer, the defined prompt's next-token distribution at that position, as it must by construction. The patches move responses the same way whichever framework's state is patched in, most often to the same response, and the all-layer patch pair recovers at most two hundredths of prompting's selectivity with quality unchanged (Table~\ref{tab:pairs}): by the reading fixed in advance, the gap is a matter of position.

\begin{table}[t]
\centering\footnotesize\setlength{\tabcolsep}{3pt}
\begin{tabular}{@{}lccc@{}}
\toprule
Model & Prompting & Pair, $\times1$ & Pair, $\times2$ \\
\midrule
\multicolumn{4}{@{}l}{\emph{Primary evaluator}} \\
Llama & 0.69 & 0.05 [$-$0.02, 0.12]$^{\dagger}$ & \textbf{0.28 [0.09, 0.46]$^{\ast}$} \\
OLMo & 0.78 & \textbf{0.17 [0.11, 0.23]$^{\ast}$} & \textbf{0.22 [0.14, 0.32]$^{\ast}$} \\
Qwen & 0.95 & \textbf{0.11 [0.04, 0.19]$^{\ast}$} & \textbf{0.28 [0.18, 0.38]$^{\ast}$} \\
\multicolumn{4}{@{}l}{\emph{Second evaluator}} \\
Llama & 0.57 & 0.05 [0.02, 0.08] & \textbf{0.27 [0.15, 0.40]$^{\ast}$} \\
OLMo & 0.68 & \textbf{0.12 [0.04, 0.21]$^{\ast}$} & \textbf{0.18 [0.09, 0.30]$^{\ast}$} \\
Qwen & 0.84 & \textbf{0.15 [0.10, 0.20]$^{\ast}$} & \textbf{0.23 [0.13, 0.34]$^{\ast}$} \\
\bottomrule
\end{tabular}
\caption{\textbf{The tone pair reaches the margin at twice the norm under both evaluators.} Tone prompting selectivity (the mean, over the two tone instructions, of each instruction\textquoteright{}s selectivity toward its own tone) and the tone pair\textquoteright{}s averaged specificity at one and two times the natural norm with its 95\% situation-bootstrap interval (10{,}000 resamples), under the primary evaluator and under the second evaluator with the same tone rubric. Unmarked: below $1/12$ by the one-sided bound; $\dagger$ undetermined; $\ast$ (bold) at or above $1/12$ with a positive lower bound.}
\label{tab:tone}
\end{table}

\paragraph{Tone calibration} Tone vectors are the mean layer-20 shift from the neutral instruction under the named optimistic and pessimistic instructions over the construction prompts (cosine between the two $0.2$ to $0.5$, against $0.85$ to $0.90$ for the framework pair). Responses are scored for an optimistic and a pessimistic outlook, each $0$ to $6$, with coherence $0$ to $2$ and refusal, by both evaluators with the same rubric; the prediction that the pair reaches $1/12$ at twice the norm was fixed before either evaluator's scores existed. Table~\ref{tab:tone} gives tone prompting selectivity and the pair's specificity under each evaluator: the two agree on the class of every pair at twice the norm and of the OLMo and Qwen pairs at the natural norm, and differ on the Llama pair at the natural norm, undetermined under the primary evaluator and positive but below $1/12$ under the second. The primary evaluator assigns full coherence to every tone response; under the second, coherence changes stay within the range of the tone instructions themselves. No response refuses, and no tone pair produces identical responses.

\end{document}